\documentclass[conference]{IEEEtran}
\IEEEoverridecommandlockouts
\usepackage{cite}
\usepackage{subcaption}
\usepackage{amsmath,amssymb,amsfonts}
\usepackage{algorithmic}
\usepackage{graphicx}
\usepackage[export]{adjustbox}
\graphicspath{ {./Images/} }
\usepackage{textcomp}
\usepackage{xcolor}
\def\BibTeX{{\rm B\kern-.05em{\sc i\kern-.025em b}\kern-.08em
    T\kern-.1667em\lower.7ex\hbox{E}\kern-.125emX}}
\begin{document}

\title{Attention on Classification for Fire Segmentation\
\thanks{}
}

\author{\IEEEauthorblockN{Milad Niknejad}
\IEEEauthorblockA{\textit{Instituto de Sistemas e Robotica, } \\
\textit{ Instituto Superior Tecnico, University of Lisbon}\\
Lisbon, Portugal \\
milad3n@gmail.com\\
}
\and
\IEEEauthorblockN{Alexandre Bernardino}
\IEEEauthorblockA{\textit{Instituto de Sistemas e Robotica, } \\
\textit{ Instituto Superior Tecnico, University of Lisbon}\\
Lisbon, Portugal \\
alex@isr.tecnico.ulisboa.pt
\\
}

}

\maketitle

\begin{abstract}
Detection and localization of fire in images and videos are important in tackling fire incidents. Although semantic segmentation methods can be used to indicate the location of pixels with fire in the images, their predictions are localized, and they often fail to consider global information of the existence of fire in the image which is implicit in the image labels. We propose a Convolutional Neural Network (CNN) for joint classification and segmentation of fire in images which improves the performance of the fire segmentation. We use a spatial self-attention mechanism to capture long-range dependency between pixels, and a new channel attention module which uses the classification probability  as an attention weight.  The network is jointly trained for both segmentation and classification, leading to improvement in the performance of the single-task image segmentation methods, and the previous methods proposed for fire segmentation. 
\end{abstract}


\section{Introduction}
Every year fire causes severe damage to persons and property all over the world. Artificial intelligence can play an important role to battle the fire incidents by early detection and localization of the fire spots.  Many methods have already been proposed for detection of fire and smoke on images and videos in different scenarios such as wildfires. Traditional methods were based on handcrafted features extracted mostly from individual  pixel colors \cite{celik2009fire, chen2004early}. Recently, analogously to many other computer vision areas, the state-of-the-art results have been achieved for fire detection using features from Convolutional Neural Networks (CNN). Methods were mainly proposed for classification of fire in images \cite{dunnings2018experimentally, barmpoutis2019fire}. Some methods consider the localization of fire in images as well \cite{zhang2016deep, chaoxia2020information}. Localization of fire is  important for determining the exact spot of the fire in images which has applications in autonomous systems and geo-referencing of the fire location. Like \cite{harkat2021fire, frizzi2021convolutional}, we consider pixel-wise segmentation for fire localization, which corresponds to the binary semantic segmentation to detect fire in images. Although bounding boxes can be used for localization, pixel-wise segmentation has advantages e.g. it can be used as input for fire propagation models. However, most segmentation methods are localized and do not consider global contextual information in images. In the case of fire detection, even recent well-known segmentation methods produce many incorrect false positive pixel segmentations for fire-like images due to these localized predictions (see Fig. \ref{fig:falsepositives}). This false positive prediction is an important issue in fire detection as it may lead to false alarms.

Recently, self-attention mechanisms have attracted lots of interest in computer vision. The  purpose of self-attention methods is to use long range information and increase the receptive field sizes of current deep neural networks \cite{wang2018non}.  Self-attention tends to capture the correlation between different image regions by computing a weighted average of all  features in different locations in which the weights are computed based on the similarities between their corresponding embeddings. Some works consider deep  architecture composed of only self-attention layers as a replacement for the convolutional networks \cite{bello2019attention}. Apart from self-attention, which uses the input features itself to compute the attention coefficients, some methods proposed to use attention based on the features extracted from other parts of the network  \cite{oktay2018attention}.

Multi-task learning methods learn simultaneously multiple correlated computer vision tasks (e.g. semantic segmentation and depth estimation) in a unified network through learning common features \cite{jafari2017analyzing, dharmasiri2017joint, zhang2018joint}. It has been shown that this multi-task learning leads to improvement in performance and reduction of training complexity compared to using separate networks for each task .  In our application,  the features in the higher layers of a CNN contain both localization and classification informations \cite{long2015fully, zhou2016learning}. Consequently, in \cite{le2019multitask}, a method is proposed for joint classification and segmentation of medical images, in which a classification network is applied to the features of the last layer of the encoder (the coarsest layer) in a encoding-decoding segmentation CNN.

In this paper, we propose a new CNN that jointly classifies and segments fire in images with improved segmentation performance. We propose an attention mechanism that uses the classification output as the channel attention coefficient of the segmentation output. This allows the overall network to consider the global classification information on the segmentation masks. Furthermore, we use a self-attention model to capture the long-range spatial correlations within each channel. Experiments show that the proposed method with the attention mechanism outperforms other methods in the segmentation metrics. It reduces the false positive results in the segmentation masks, while at the same time, is able to identify small scale fires in images, resulting in state-of-the-art results among fire segmentation methods. 

In the following sections, we first mention related works for segmentation, multi-task learning, and self-attention. We then describe our proposed method in detail, and finally compare our method with other segmentation, and multitask classification-segmentation methods.

\section{Related works}

Traditional methods for fire detection mainly use hand-crafted features such as color features  \cite{celik2009fire}, \cite{chen2004early}, covariance-based features \cite{habibouglu2012covariance}, wavelet coefficients \cite{toreyin2006computer}, and then classify the obtained features using a vector classifier e.g. a Support Vector Machine (SVM). Recently, methods based on CNN have improved the performance of fire and smoke detection noticeably. The method in \cite{dunnings2018experimentally} uses simplified structures of the Alexnet \cite{krizhevsky2012imagenet} and Inception networks \cite{szegedy2015going} for fire image classification. In \cite{barmpoutis2019fire}, Faster Region-CNN (R-CNN) \cite{ren2015faster} is used to extract fire candidate regions, and is further processed by multidimensional texture analysis using Linear Dynamical Systems (LDS) to classify the fire images. 

Some works  consider localization of fire in images, beyond classification. In \cite{zhang2016deep}, a method for classification and patch-wise localization was proposed in which the last convolutional layer of the classification network is used for the patch classification. In \cite{chaoxia2020information}, a combination of color features, and Faster R-CNN is used to increase the efficiency of the algorithm by disregarding some anchors of R-CNN based on some color features. Some methods consider pixel-wise segmentation for fire images as well. In \cite{harkat2021fire}, deep-lab semantic segmentation is adapted for pixel segmentation of fire. In \cite{frizzi2021convolutional}, a new CNN architecture is proposed for segmentation of fire in images.

In computer vision, single end-to-end multi-task networks have shown promising results for the tasks that have cross-dependency such as semantic segmentation and depth estimation \cite{jafari2017analyzing, dharmasiri2017joint}. They benefit from learning common features. It has been shown that exploiting the cross-dependency between the tasks lead to to improvement in performance compared to the networks independently trained for the two tasks \cite{dharmasiri2017joint}. It has other benefits such as reducing the training time. It is known that the features in the last convolutional layer in CNNs trained for classification have also spatial information for localization \cite{long2015fully, zhou2016learning}. Le et. al. \cite{le2019multitask} proposed a method for joint classification and segmentation for cancer diagnosis in mammography, in which the last convolution layer of the encoder in the segmentation network is used for the global classification. 

Self-attention models have recently demonstrated improved results in many computer vision tasks \cite{wang2018non, bello2019attention, fu2019dual}. Self-attention models compute attention coefficients based on the similarities between input features. New features are then obtained by a weighted average of the input features with the self-attention coefficients. Beyond self-attention, there are also attention mechanisms proposed for image classification \cite{wang2017residual, jetley2018learn}, and semantic segmentation \cite{oktay2018attention}, in which the attention weights are computed using the features in other parts of the CNN.  
 
 \section{Proposed method}
 A simple approach to learn a joint classification and segmentation in a unified CNN is to classify images based on the features after global pooling of the coarsest layer (last encoding layer) of the encoder-decoder segmentation network. The network can be jointly trained with the classification and segmentation labels through a weighted loss. This approach has been previously proposed in \cite{le2019multitask}  for medical imaging application. 
 
 \begin{figure*}
 \centering
\includegraphics[width=.8\textwidth, frame]{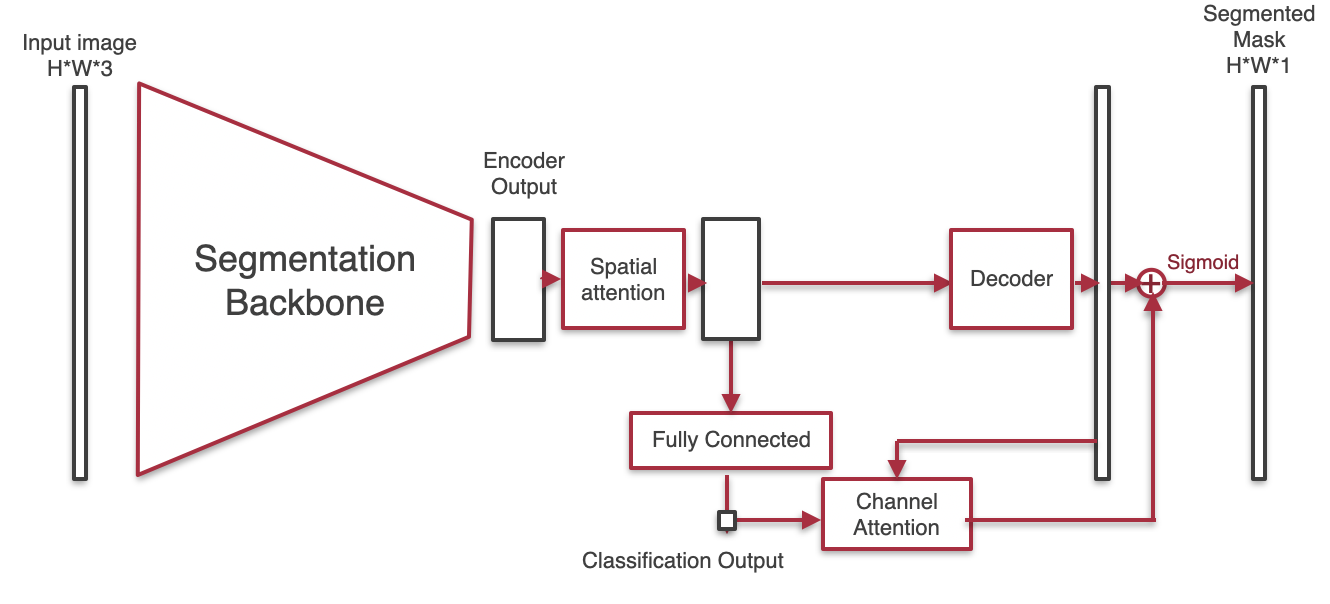}
\caption{Proposed CNN architecture for joint classification and segmentation; for the segmentation backbone deeplab v3 \cite{chen2017rethinking} is used. }
\label{fig:network}
\end{figure*}

In this paper we add two attention modules to consider both global classification score and correlation in the spatial locations, in the segmentation predictions. As the first attention module, we propose to use a channel attention in the segmented output. As shown in Fig. \ref{fig:network}, it multiplies the attention weight to the output channel, in which the weight is the probability assigned by the classification branch of the network. The channel is then added to the resulting features, similar to self-attention approaches \cite{wang2018non}. Let, $\mathbf{x} \in \mathbb{R}^{W \times H \times 3}$ be the RGB image, and  let $s(\mathbf{x}) \in \mathbb{R}$ and $\mathbf{A}(\mathbf{x}) \in \mathbb{R}^{W \times H \times 1}$ be the classification probability and segmentation features extracted by the CNN, respectively. $\mathbf{A}(\mathbf{x})$ could be the output of any segmentation network. In this paper, we use deeplab v3+ encoder and decoder \cite{chen2018encoder} to extract the features. $s(\mathbf{x}) \in [0,1]$ is computed by the sigmoid function of the classification scores obtained by the classification branch (see Fig. \ref{fig:network}).  Following \cite{fu2019dual}, we compute the channel attention model as
 \begin{equation}
\mathbf{A}'({\mathbf{x}})=\mathbf{A}(\mathbf{x})+\alpha s(\mathbf{x}) \mathbf{A}(\mathbf{x})
 \label{eq:module}
 \end{equation}
 where $\mathbf{A}'$ indicates the features after applying the attention module, and $\alpha$ is a parameter which is initialized to zero and learnt during training \cite{fu2019dual}. The method  in \cite{fu2019dual}  used a channel attention model for the segmentation in which the channel weights are obtained by the features themselves  using a self-attention approach. However in our method, the weight $s(\mathbf{x})$ is the classification probability which is computed in the classification branch. This approach is supposed to reduce the false positive results as the correct classification output $s(\mathbf{x})$ for a non-fire image  is close to zero, so it attenuates the activation of the  segmentation output $\mathbf{A}'$. In the case of fire, a value of $s(\mathbf{x})$ close to one helps to recognize even small portions fire in images. This also encourages the consistency of the results between the segmentation and classification outputs.
 
 \begin{figure}
 \centering
\includegraphics[width=.45\textwidth, frame]{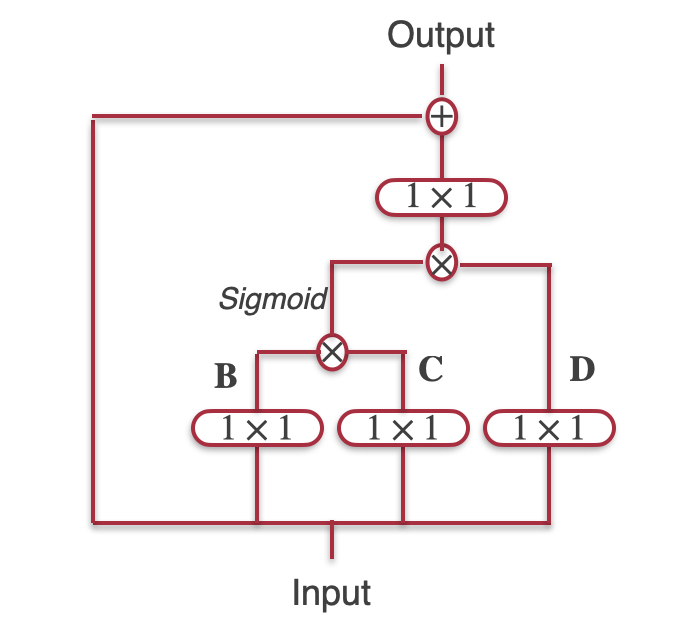}
\caption{Spatial self-attention module used on our method to capture long range spatial dependency which is proposed in \cite{wang2018non}. The rounded boxes indicate the convolution operator. }
\label{fig:att}
\end{figure}
 
 Although the above global attention scheme considers the image label information, the performance of the method can be further improved by considering the correlation between the features. To achieve this, besides the classification attention model, we also apply spatial attention model to consider the correlation between features in different locations within each channel. This attention model is exactly the same as proposed in  non-local neural networks of \cite{wang2018non}, in which each feature is replaced by  a weighted average of all features based on the similarities between their corresponding embeddings. The general structure of the spatial attention module as shown in Fig. \ref{fig:att}  is to apply two $1\times1$ convolution layers to the input features, and reshape the result to obtain the similarity embeddings $\mathbf{B} \in \mathbb{R}^{N \times C}$ and $\mathbf{C} \in \mathbb{R}^{N \times C}$, where $N=H \times W$, and $C$ is the number of channels. The two matrices are used to compute a similarity matrix $\mathbf{S}=\mathbf{B}\mathbf{C}^T$. The row-wise softmax of the matrix $\mathbf{S}$ is multiplied to the matrix $\mathbf{D}$, which results from another $1 \times 1$ convolution to the input. The resulting is added to the input. Figure \ref{fig:att} shows the self-attention module in our method.
 
 Spatial and channel attentions have been used for segmentation of general images in \cite{fu2019dual}. However, our method uses completely different channel attention module based on the classification probabilities, while \cite{fu2019dual} uses a self-attention module.

 Following the common approach in multitask learning, we use a weighted sum of the classification and segmentation losses. Let $L_S$, and $L_C$, be the segmentation and classification losses, respectively. The training loss is computed by
 \begin{equation}
 L=\lambda L_S+ (1-\lambda) L_C
 \end{equation}
 where $\lambda \in [0, 1]$ is an appropriate regularization parameter. We use binary cross-entropy loss for both $L_C$, and $L_S$.
 
 \section{Experimental results}
 In this section, We evaluate our proposed method and compare it to other segmentation methods and multitask methods for joint segmentation and classification. In order to evaluate the performance for false positive segmentation, we compute the label inferred from the segmented image by $\mathbf{1}(\sum_{i,j} \mathbf{M}_{i,j})$ where $ \mathbf{M}_{i,j}$ indicates the output mask at pixel $i,j$, and $\mathbf{1}$ is the indicator function. It is considered zero if all pixels in the output mask are zero, and one otherwise. We use the accuracy between the segmented label and the image label in our comparisons which is called average consistency in Table \ref{tab:results}.

We create a dataset by combining RGB images and their associated segmentation masks in the Corsican fire dataset \cite{toulouse2017computer}, and non-fire images in \cite{chino2015bowfire}, containing some images which are likely to cause false positive results. We divided the dataset into train, validation, and test groups with 60, 20 and 20 percent, respectively.

\begin{figure*}
\centering
\begin{subfigure}[b]{0.18\textwidth} \includegraphics[width=\textwidth]{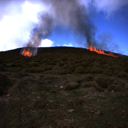} \end{subfigure}
\begin{subfigure}[b]{0.18\textwidth} \includegraphics[width=\textwidth]{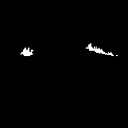} \end{subfigure}
\begin{subfigure}[b]{0.18\textwidth} \includegraphics[width=\textwidth]{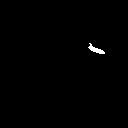}\end{subfigure}
\begin{subfigure}[b]{0.18\textwidth} \includegraphics[width=\textwidth]{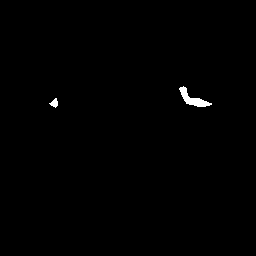}\end{subfigure}
\begin{subfigure}[b]{0.18\textwidth} \includegraphics[width=\textwidth]{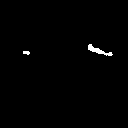}\end{subfigure}

\medskip

\begin{subfigure}[b]{0.18\textwidth} \includegraphics[width=\textwidth, height=\textwidth]{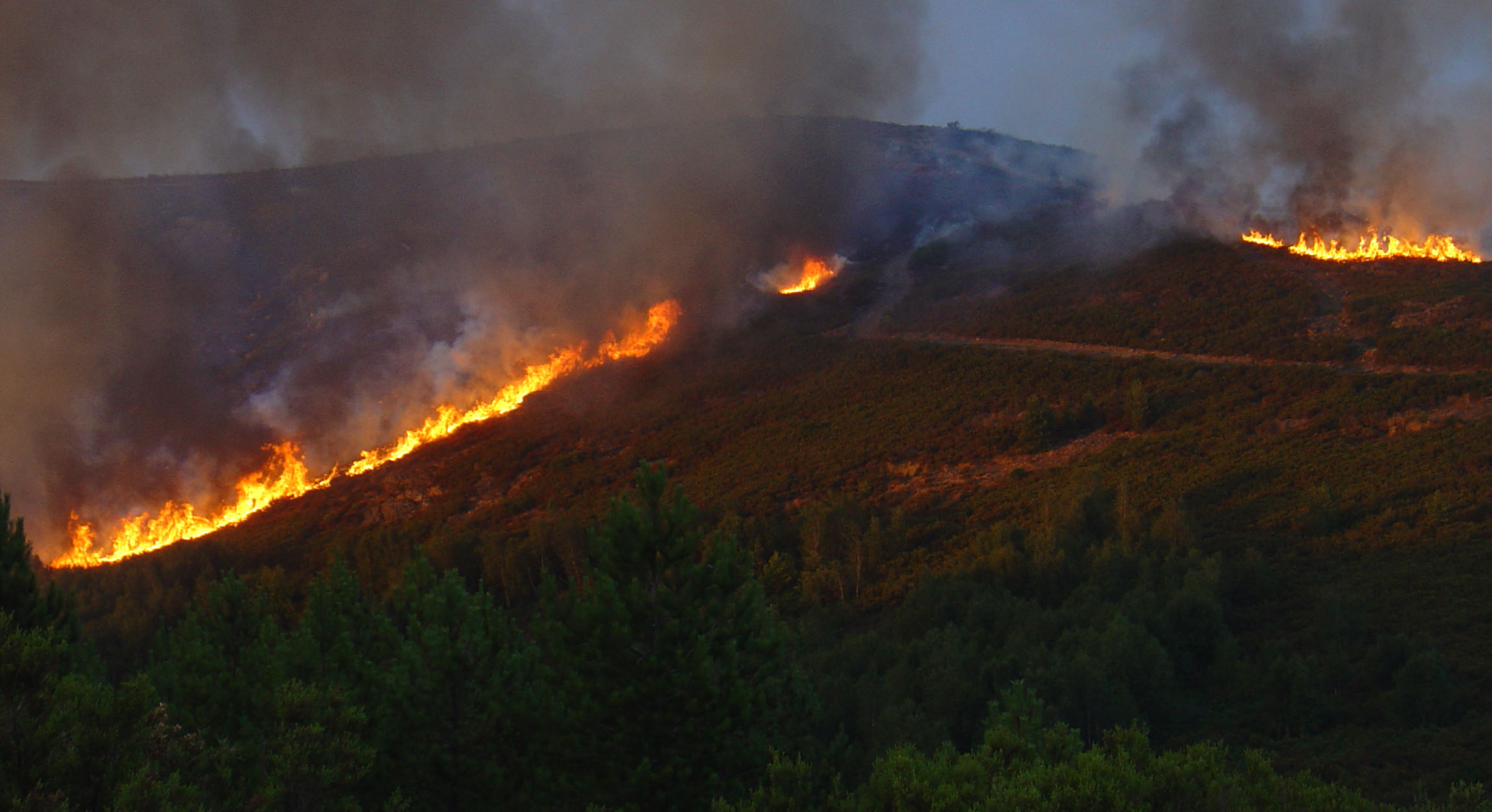} \end{subfigure}
\begin{subfigure}[b]{0.18\textwidth} \includegraphics[width=\textwidth, height=\textwidth]{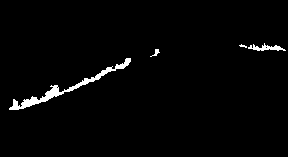}\end{subfigure}
\begin{subfigure}[b]{0.18\textwidth} \includegraphics[width=\textwidth]{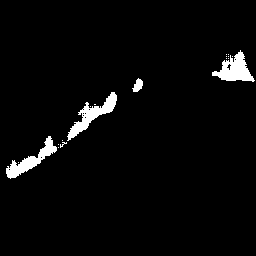} \end{subfigure}
\begin{subfigure}[b]{0.18\textwidth} \includegraphics[width=\textwidth]{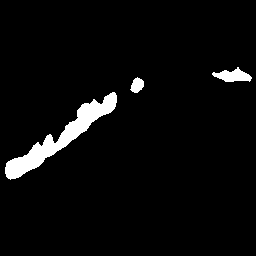}\end{subfigure}
\begin{subfigure}[b]{0.18\textwidth} \includegraphics[width=\textwidth]{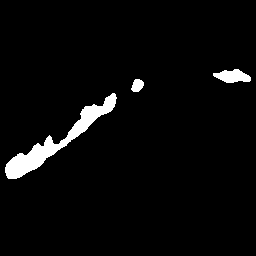}\end{subfigure}

\medskip
\begin{subfigure}[b]{0.18\textwidth} \includegraphics[width=\textwidth]{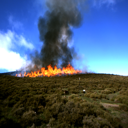} \caption{Original image \\ \hfill} \end{subfigure}
\begin{subfigure}[b]{0.18\textwidth} \includegraphics[width=\textwidth]{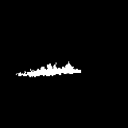} \caption{ground-truth \\}\end{subfigure}
\begin{subfigure}[b]{0.18\textwidth} \includegraphics[width=\textwidth]{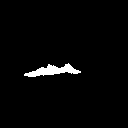}\caption{U-net}\end{subfigure}
\begin{subfigure}[b]{0.18\textwidth} \includegraphics[width=\textwidth]{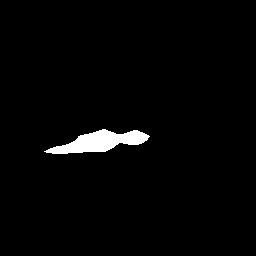}\caption{Deeplab \\}\end{subfigure}
\begin{subfigure}[b]{0.18\textwidth} \includegraphics[width=\textwidth]{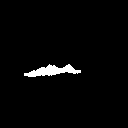}\caption{Proposed }\end{subfigure}

\caption{Examples of the segmentation of fire in images in our proposed method compared to other methods.}
\label{fig:segmented}
\end{figure*}

\begin{figure*}
\centering
\begin{subfigure}[b]{0.2\textwidth} \includegraphics[width=\textwidth, height=\textwidth]{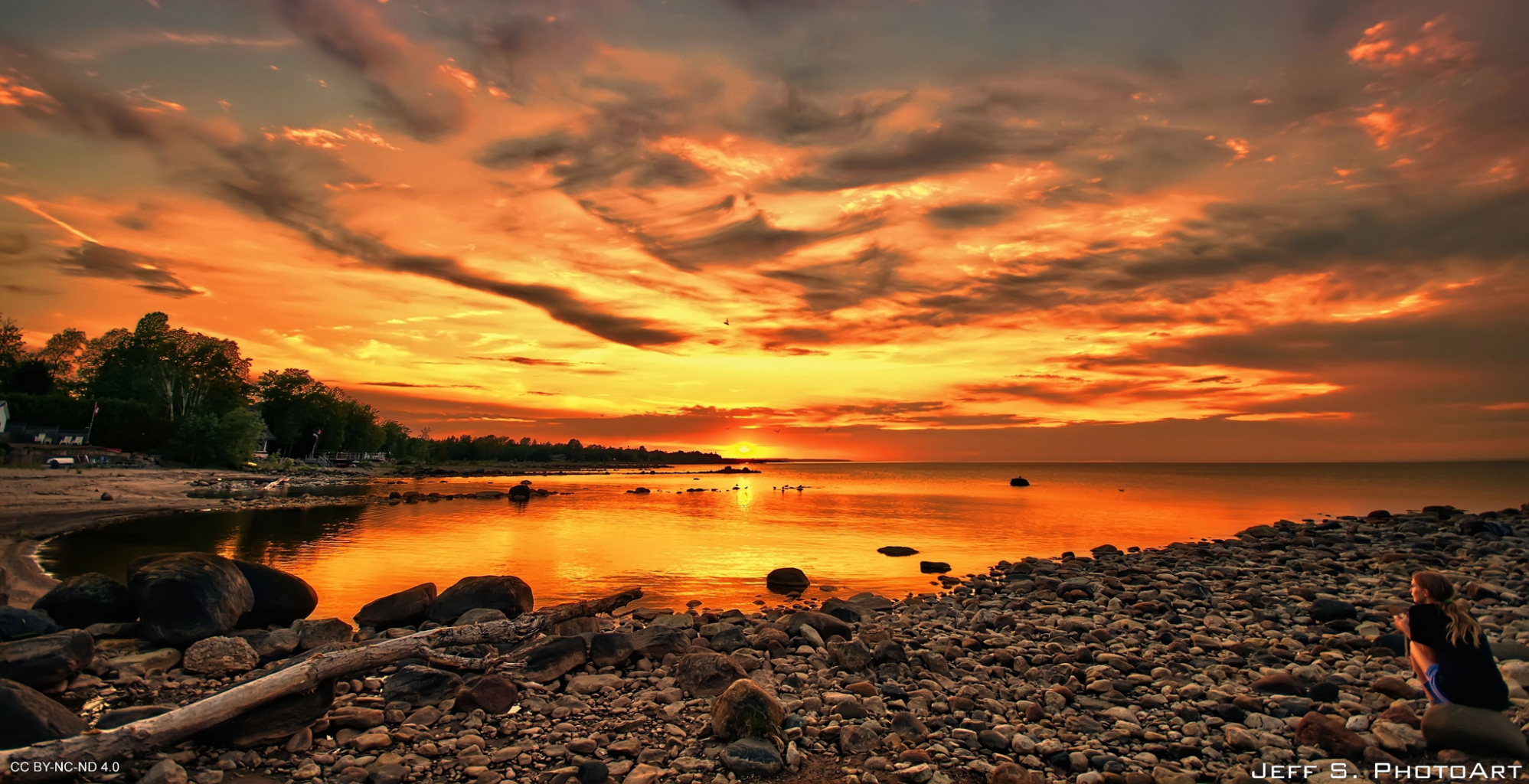} {\\} \end{subfigure}
\begin{subfigure}[b]{0.2\textwidth} \includegraphics[width=\textwidth]{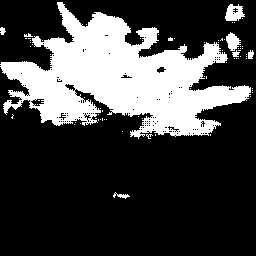}{\\}\end{subfigure}
\begin{subfigure}[b]{0.2\textwidth} \includegraphics[width=\textwidth]{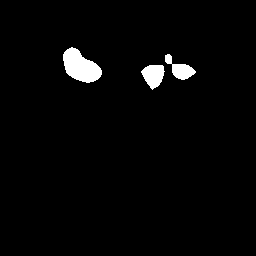}{\\}\end{subfigure}
\begin{subfigure}[b]{0.2\textwidth} \includegraphics[width=\textwidth]{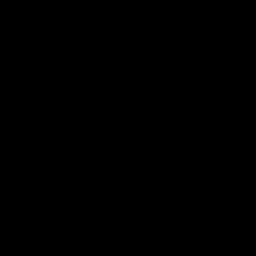} {\\} \end{subfigure}

\begin{subfigure}[b]{0.2\textwidth} \includegraphics[width=\textwidth, height=\textwidth]{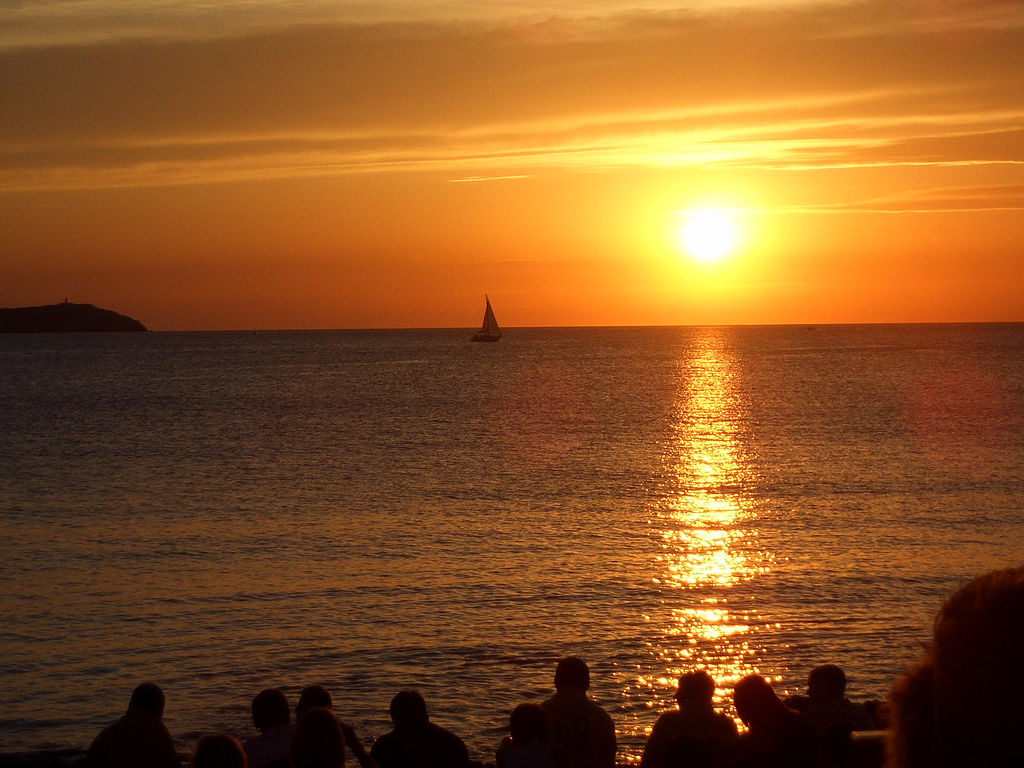} {\\ Original image \hfill} \end{subfigure}
\begin{subfigure}[b]{0.2\textwidth} \includegraphics[width=\textwidth]{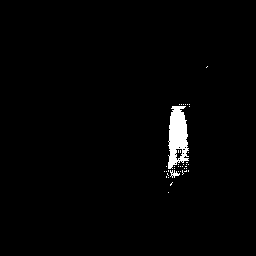}{\\ U-net \cite{ronneberger2015u}}\end{subfigure}
\begin{subfigure}[b]{0.2\textwidth} \includegraphics[width=\textwidth]{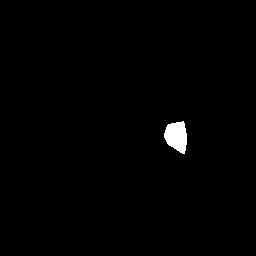}{\\ Deeplab \cite{harkat2021fire}}\end{subfigure}
\begin{subfigure}[b]{0.2\textwidth} \includegraphics[width=\textwidth]{deep_lab_att_not_fire002} {\\ Proposed} \end{subfigure}

\caption{Examples of segmentation of fire part for images that are likely to produce false positive results.}
\label{fig:falsepositives}
\end{figure*}

\begin{table*}[]
\centering
\begin{tabular}{|l|l|l|l|l|l|}
\hline
         & \multicolumn{1}{l|}{Classification metrics} & \multicolumn{2}{l|}{Segmentation Metrics} & Consistency    \\ \hline
         & Accuracy                  & mean Accuracy            & mean IOU            & Avg. Consistency    \\ \hline
U-net \cite{ronneberger2015u}     & -                                & 96.88               & 87.45               & .8521          \\ \hline
Deeplab \cite{harkat2021fire} & .-                       & 97.18               & 88.34               & .8876          \\ \hline
Fire segmentation in \cite{frizzi2021convolutional}& -                          & 97.06               & 88.02              & .8623          \\ \hline
multi-task network in \cite{le2019multitask}& {98.75}               & {96.55}      & {87.22}      & {.8912} \\ \hline
Naive multi-task& {98.75}             & {97.21}      & {90.02}      & {.9654} \\ \hline
Proposed& {99.12}               & \textbf{98.02}      & \textbf{92.53}      & \textbf{.9823} \\ \hline
\end{tabular}
\caption{Comparison of the proposed method with the baseline for classification-segmentation, and U-net for segmentation.}
\label{tab:results}
\end{table*}

In this section, our proposed segmentation method is compared to U-net  \cite{ronneberger2015u}, deeplab adapted for fire segmentation \cite{harkat2021fire}, and the method in \cite{frizzi2021convolutional} which proposed a new architecture for fire  segmentation. We also compare our proposed method with other joint classification-segmentation methods. Inspired by \cite{le2019multitask}, we consider a multi-task approach, which applies a classification network to the output of the encoder of the segmentation network. This method corresponds to our proposed method in which all attention blocks are removed.   We also consider a simple approach for removing false positives in segmentation in which the segmentation mask is set to zero if the classification output is zero (without using any attention). This basically relies on the classification output for segmentation. We call this method the naive approach.

The proposed method is implemented in the following settings. The encoder of Deeplab-v3+ is used as the encoder backbone for the segmentation network. The  network is then initialized by pre-trained ImageNet weights for the segmentation backbone, and i.i.d normal random weights with mean zero and standard deviation of $.05$ for the classification branch. The weights are learned during the training by the ADAM algorithm \cite{kingma2014adam} with initial learning rate of $5 \times 10^{-4}$, and a weight decay of $10^{-5}$. The loss regularization parameter $\lambda$ is set to $.6$, empirically, to achieve the best performance in terms of the overall validation loss. All other methods were trained in our dataset with ADAM algorithm with the parameters which performs the best in the validation set. For fire segmentation method in \cite{frizzi2021convolutional}, we trained our own implementation as the source codes are not available online.

We report the result of our proposed method and other mentioned methods, on the test set, in Table \ref{tab:results}. The main metric for assessing the performance of semantic segmentation methods is intersection over union (IOU). This value is computed in this table for the classes of background and fire. In the classification metric, we compare the accuracy between the ground truth image labels and the predicted labels. This metric is obviously valid for multitask networks that have the image classification branch. In the segmentation metrics, the pixel accuracy (averaged over all tests images) and mean IOU are reported. 
 As it can be seen, in the IOU segmentation metric, the proposed method outperforms segmentation methods of U-net, Deeplab adapted for fire segmentation in \cite{harkat2021fire}, and the fire segmentation method in \cite{frizzi2021convolutional}. IOU  is also improved over the joint segmentation and classification method of \cite{le2019multitask}, and the naive approach described above. 
 Some examples of fire segmentation on test dataset are shown in Fig. \ref{fig:segmented}. In the first row, it can be seen that our method could capture small portion of fire in the image. 
 Besides that, based on the results on the table, the proposed method performs better in the consistency metric which is defined in the first paragraph of this section (the accuracy between the inferred label from the segmentation output and the ground truth label). This metric shows that the segmented image better corresponds to the image label in our method, i.e. reducing the cases in which some pixels are assigned as fire in non-fire images. This is a common problem in fire segmentation as illustrated in Fig. \ref{fig:falsepositives} for two images which are prone to false positive outputs. As it can be seen, other methods  mistakenly select some parts of both images as fire, while the proposed method correctly does not segment any pixel as fire.

\section{Conclusion}
In this paper, we proposed a method for joint classification and segmentation of fire in  images based on CNN using attention. We used a channel attention mechanism in which the weight is based on the  output in the classification branch. A self-attention mechanism is used for spatial attention. Our method shows improved segmentation results over other fire segmentation methods, and other multitask CNN structures.

 \bibliography{ref} 
\bibliographystyle{IEEEtran}

\end{document}